# Accurate and robust image superresolution by neural processing of local image representations

Carlos Miravet[1,2] and Francisco B. Rodríguez[1]

[1] Grupo de Neurocomputación Biológica (GNB), Escuela Politécnica Superior,
Universidad Autónoma de Madrid, 28049 Madrid, Spain
carlos.miravet@uam.es, f.rodriguez@uam.es
[2] SENER Ingeniería y Sistemas, S. A.,Severo Ochoa 4 (P.T.M.), 28760 Madrid, Spain

**Abstract.** Image superresolution involves the processing of an image sequence to generate a still image with higher resolution. Classical approaches, such as bayesian MAP methods, require iterative minimization procedures, with high computational costs. Recently, the authors proposed a method to tackle this problem, based on the use of a hybrid MLP-PNN architecture. In this paper, we present a novel superresolution method, based on an evolution of this concept, to incorporate the use of local image models. A neural processing stage receives as input the value of model coefficients on local windows. The data dimensionality is firstly reduced by application of PCA. An MLP, trained on synthetic sequences with various amounts of noise, estimates the high-resolution image data. The effect of varying the dimension of the network input space is examined, showing a complex, structured behavior. Quantitative results are presented showing the accuracy and robustness of the proposed method.

## Introduction

Image superresolution [1, 2] involves the processing of an image sequence to generate a high-resolution description of the underlying scene. From the earliest algorithm proposed by Tsai and Huang [3], a number of approaches have been proposed. Of these, bayesian MAP (Maximum A Posteriori) methods [4,5] have gained particular acceptance due to their robustness and their capability to incorporate *a priori* constraints. The main drawback of these methods comes from their associated high computational loads, as they use iterative techniques in spaces of high dimensionality.

Recently [6, 7], the authors have proposed a neural network based technique that provides results comparable to classical methods with a substantial decrease in computational complexity. This technique estimates image values in a dense grid using an irregular interpolation scheme, with distance dependent interpolation weights. Optimal distance-to-weight mappings are learned from synthetic sequences and corresponding high-resolution images, using a hybrid MLP-PNN (Multi Layer Perceptron – Probabilistic Neural Network) architecture. In a second step, high-resolution image values are restored from estimated grid values using optimal filters.

The use of interpolation schemes based on distance dependent weights, no matter how optimally these weights can be tuned, poses a fundamental limit on the attainable performance as, being independent of local image structure, the method is forced to operate in the same way near an image edge or inside a uniform image patch. In this paper, an evolution of our previous distance-based algorithm is presented, which makes explicit use of local image representations. As in our previous approach, the proposed system learns from examples how to perform superresolution. In this way, the computational load is mostly displaced to the off-line learning process, enabling a fast, non-iterative response in the network deployment phase.

The proposed method is based on the sequential application of two processing steps. In the first step, sequence pixels are projected onto the high-resolution frame, and local image representations are built for each site of an embedded high-resolution grid. In the second step, the image representation coefficients in the neighborhood of each grid site are processed by a neural network to estimate the high-resolution image values. The dimensionality of the network input data is previously reduced by application of a PCA (Principal Component Analysis) technique.

Out method has shown to provide excellent results over a wide range of input noise levels. In the following sections, we detail the processing steps involved, and present experimental results that include a quantitative comparison of several methods. In the last section, a brief discussion of the main results obtained is presented.

## Local image representation

The first step of our superresolution method computes a local image representation for each site of the high-resolution (HR) grid to be estimated. These local representations are built using the sequence pixels values projected onto the HR grid. The projection operation requires the previous knowledge of the geometrical transformations that relate input sequence frames. To estimate this data, an adaptation of a classical sub-pixel registration procedure [8] has been used.

**Table 1.** RMS errors for different interpolation schemes.

| Method | $\sigma=0$ | $\sigma=5$ | $\sigma=10$ | $\sigma=20$ |
|---|---|---|---|---|
| NN_SEQ | 9.28 | 10.40 | 13.48 | 21.97 |
| **Distance-based interpolation** | | | | |
| Inverse distance weight | 6.82 | 7.48 | 9.23 | 14.16 |
| MLP-PNN | 5.61 | 6.00 | 6.85 | 8.43 |
| **Polynomial Models** | | | | |
| Order 1 | 7.24 | 7.37 | 7.51 | 8.48 |
| Order 2 | 2.90 | 3.56 | 5.04 | 8.75 |
| Order 3 | 2.34 | 3.29 | 5.25 | 9.72 |

Polynomial models have been used to describe the local image structure at sub-pixel level. Polynomial coefficients are determined by a general linear least squares technique, with matrix inversion performed using singular value decomposition, SVD [9]. The use of SVD improves robustness when the problem is close to singular, due, for instance, to an inadequate model order selection in an image patch, or induced by a large image noise level.

In table 1 are presented the root mean squared (RMS) errors for different irregular interpolation methods. Polynomial models with orders ranging from 1 to 3 have been considered. For comparison, the error corresponding to the direct selection of the nearest pixel (SEQ_NN) has been included, together with two distance-based interpolators: the first processing step of our previous method (MLP-PNN), and a method recently proposed in the literature [10], which makes use of inverse distance weights. As it can be observed, interpolation with second order local polynomials provides distinct advantages over distance-based methods at low and medium noise levels. Increasing polynomial order over this point result only in marginal gains in performance. In view of these results, a second order polynomial has been considered in our subsequent work.

**Dimensionality reduction by PCA. Eigenimages of coefficients**

The dimension of network input space is given by the model dimension (6 for a second order polynomial) times the size of the considered local neighborhood. Principal component analysis (PCA) has been applied to reduce the dimensionality of this space. PCA [11] is a linear technique that yields minimal representational error (in terms of mean squared error, MSE) for a given reduction in the dimensionality space. PCA operates by projecting the input data onto an orthogonal basis of the desired dimension, where the basis vectors are the eigenvectors of the input data covariance matrix, ranked in order of decreasing eigenvalues.

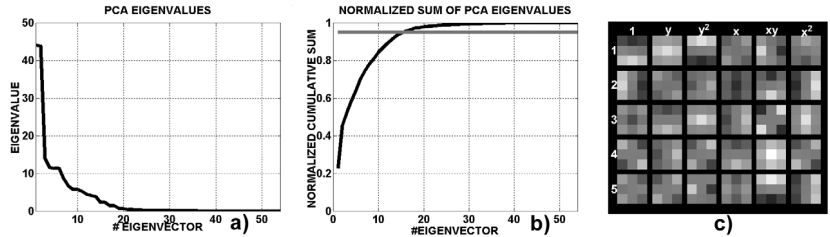

**Fig. 1.** Results of the PCA process, for a 3x3 local neighborhood: a) eigenvalues sorted in decreasing order; b) percentage of input variance maintained after dimensionality reduction, as a function of the number of used eigenvectors; c) first five eigenimages.

In figure 1 a, b) are presented, for a 3x3 local neighborhood, the obtained eigenvalues sorted in decreasing order, and the percentage of the input variance represented after dimensionality reduction, as a function of the number of used eigenvectors. Using a common heuristic approach, such as maintaining enough components to explain 95% of the input variance, will lead to the selection of the first 16 eigenvectors to build the projection basis. In the next section we will see that heuristics of this type are not appropriate for this problem, as use of low variance eigenvectors can have a considerable impact on the final prediction error.

In figure 1 c) are presented the five eigenvectors of highest eigenvalue, with components rearranged in the form of eigenimages. The preliminary experiments conducted show their stability under variations of input noise and image content, defining basic patterns of spatial variation of local polynomial models in images examined at sub-pixel level. The projection of the local image models on this set of spatial patterns is used as input data to the network, as described in the next paragraph.

**Neural network training results**

To estimate the HR pixels, it has been used a multi-layer perceptron (MLP) architecture [11], with two layers of neurons. The hidden layer is composed of 10 neurons with hyperbolic tangent activation functions. These neurons are connected to a single neuron in the output layer, with a linear activation function. The output of this neuron provides a zero-mean, unit-variance estimation of the high-resolution central pixel. Renormalization provides the final value. A diagram of the system is presented in figure 2 a).

The network has been trained using a conjugate gradient descent method on synthetic data sets with various noise levels. The training data has been generated synthetically from a set of 23 high-resolution images of urban scenes acquired by the *Quickbird* satellite imaging system.

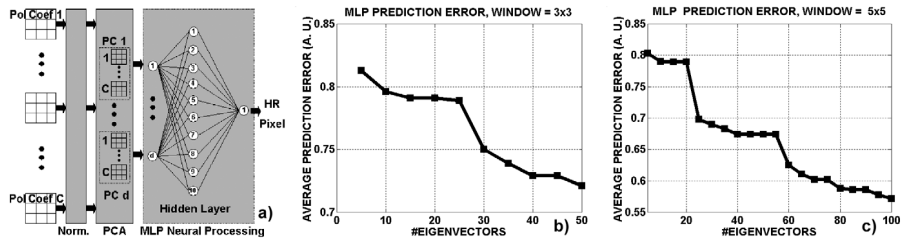

**Fig. 2.** a) Diagram of the superresolution neural processing stage; b) network average prediction error, measured on a validation test set, as a function of the dimensionality of the network input space for a 3x3 neighborhood; c) average prediction error for a 5x5 neighborhood.

The neural system has been trained for several input sizes, to study quantitatively the effect on the predicting error of augmenting the input data with the value corresponding to each sorted principal component. In figure 2 is presented the evolution of this error as a function of the number of eigenvectors retained in the dimensionality reduction step, for local windows of sizes 3x3 and 5x5. As it is apparent, the error decreases non-gradually, with almost flat zones interleaved with step decreases in the prediction error. Furthermore, these step changes occurred also in zones where the representational error of the dimensionality reduction step might be seen in principle as negligible (see figure 1, for a 3x3 window), highlighting the inadequacy of an MSE-based criteria in this case to select the dimension of the network input space.

**Experimental Results**

The accuracy and stability of the proposed method has been tested with both synthetic and outdoor sequences of different image content. Here, we present the results of applying superresolution on synthetic image sequences with various noise levels, to enable the quantitative comparison, in terms of RMS, of the prediction error of different methods. The sequences have been generated from a high-resolution image of an urban area acquired with the IKONOS satellite imaging system. This image has not been used during the training process. From this image, two sequences of 25 images (corresponding to 1s in the CCIR standard) were generated, one corresponding to the noiseless case, and the second one corrupted highly by addition of Gaussian noise (σ = 20 gray levels).

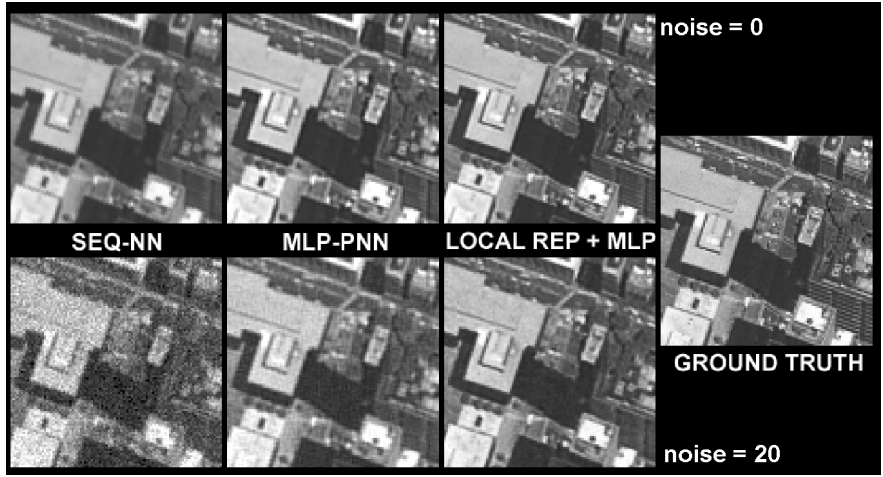

**Fig. 3.** Superresolution results, for a scaling factor of 2, on a synthetic sequence of an urban area using different methods. In the first row, are presented the results for a noiseless input sequence. The second row contains the results for a sequence corrupted with Gaussian noise of σ = 20 gray levels. Each column contains the results of a different method. The ground truth is presented at the right of the image.

On figure 3 are presented the results obtained with three superresolution methods. In the first column (SEQ-NN) are shown the results obtained estimating the HR pixels as the value of the nearest projected sequence pixel. This method produces jagged edges in the noiseless case, and no input data noise reduction. Both effects are reduced, as can be seen in the second column, when using the MLP-PNN technique, with networks specialized in the appropriate noise range. Finally, the use of the proposed method, with 3x3 neighborhoods and 40 eigenvectors, improves both image definition and noise rejection at all input noise levels, as can be seen in the third column of the figure. These perceptual results are supported by the RMS error figures reported in table 2.

**Table 2.** High-resolution RMS reconstruction errors for several methods

|  | ZOOM | SEQ-NN | MLP-PNN | LOCALREP+MLP |
|---|---|---|---|---|
| noiseless sequence (σ=0) | 19.39 | 13.04 | 11.24 | 8.28 |
| noisy sequence (σ=20) | 24.08 | 23.58 | 14.60 | 11.06 |

## Discussion

A superresolution method based on neural processing of local image representations has been developed. The method process representation coefficients on local windows to estimate the HR image values. The dimensionality of input data to this network is firstly reduced by application of a PCA technique. The eigenimages obtained have been shown to be stable for a wide range of noise levels and image contents, representing intrinsic properties of image structure at sub-pixel levels. Variations of prediction error with input size have been examined, showing a non-gradual, structured behavior, reflecting the inadequacy of MSE heuristics in this problem. The relation of the results with those provided by non-linear methods, such as ICA [12], will be investigated in future work. The experimental results obtained show the accuracy and robustness of the developed method.

**Acknowledgments**
This work has been partially supported by grants BFI2003-07276 and TIN2004-04363-C03-03, and by PROFIT project FIT-330100-2004-91.

## References

1. S. Borman, R. Stevenson, "Super-resolution from image sequences-a review", *Midwest Symposium on Circuits and Systems* (1998).
2. S. C. Park, M. K. Park, M. G. Kang, "Super-resolution image reconstruction: a technical overview", *IEEE Signal Processing Magazine*, 21-35 (2003).
3. R.Y.Tsai and T.S.Huang (Ed.), "Multiframe image restoration and registration", in *Advances in Computer Vision and Image Processing* volume 1, pages 317-339, JAI Press Inc.(1984).
4. R. C. Hardie, K. J. Barnard, J. G. Bognar, E. E. Armstrong, and E. A. Watson, "High-resolution image reconstruction from a sequence of rotated and translated frames and its application to an infrared imaging system", *Optical Engineering*, **37**(1), 247-260 (1998).
5. R. R. Schultz and R.L. Stevenson. "Extraction of high-resolution frames from video sequences", *IEEE Trans. Image Processing*, vol. 5, nº 6, pp. 996-1011 (1996).
6. C. Miravet and F. B. Rodríguez, "A hybrid MLP-PNN architecture for fast image super-resolution", Joint 13th International Conference on Artificial Neural Networks / 10th International Conference on Neural Information Processing (ICANN/ICONIP), *Lecture notes in Computer Science*, vol. 2714, Springer-Verlag, pp. 417-424 (2003).
7. C. Miravet and F. B. Rodríguez, "A two-step neural network based algorithm for fast high-resolution image reconstruction", *Image and Vision Computing* (submitted).
8. M.Irani, S.Peleg, "Improving resolution by image registration", CVGIP: *Graphical Models and Image Processing*. 53, 231-239 (1991).
9. W. Press, S. Teukolsky, W. Vetterling, B. Flannery, *Numerical recipes in C*, Cambridge University Press, 2º Ed. (1992)
10. M. S. Alam, J. G. Bognar, R. C. Hardie and B. J. Yasuada, "Infrared image registration and high-resolution reconstruction using multiple translationally shifted aliased video frames", *IEEE Transactions on instrumentation and measurement*, vol. 49, nº 5 (2000).
11. A.Bishop, *Neural Networks for Pattern Recognition*. Oxford University Press (1995).
12. A. Hyvärinen, J. Karhunen, E. Oja, *Independent Component Analysis*. Wiley-InterScience (2001).